\documentclass[letterpaper, 10 pt, conference]{ieeeconf}  % Comment this line out if you need a4paper

\IEEEoverridecommandlockouts                              % This command is only needed if 
\usepackage{times}
\usepackage[pdftex]{graphicx}
\usepackage{amsmath,amssymb,amsopn,amstext,amsfonts}
\usepackage{cancel}
\usepackage{cite}
\usepackage{pdfsync}
\usepackage{balance}
\usepackage{color}
\usepackage{mathtools}
\usepackage[ruled,vlined,linesnumbered]{algorithm2e}
\usepackage{bm}

\usepackage{diagbox}
\usepackage{float}
\usepackage{epstopdf}
\usepackage{pifont}
\usepackage{fixltx2e}
\usepackage{multirow}
\usepackage{url}
\usepackage{svg}
\usepackage[linkcolor=black,citecolor=black,urlcolor=black,colorlinks=true,bookmarks=true]{hyperref}

\usepackage{algpseudocode}
\usepackage{booktabs}
\usepackage{caption}
\usepackage{multicol}
\usepackage{verbatim}

\usepackage{mathrsfs} 
\usepackage{diagbox}
\usepackage{subcaption}

\graphicspath{{./figures/}}
\DeclareGraphicsExtensions{.png,.jpg,.eps,.pdf}
\IEEEoverridecommandlockouts
\makeatletter
\def\endthebibliography{%
	\def\@noitemerr{\@latex@warning{Empty `thebibliography' environment}}%
	\endlist
}

\title{\LARGE \bf
Primitive-Planner: An Ultra Lightweight Quadrotor Planner with Time-optimal Primitives
}

\author{Jialiang Hou$^{1, 3}$, Neng Pan$^{2, 3}$, Zhepei Wang$^{2, 3}$, Jialin Ji$^{2, 3}$, Yuxiang Guan$^{1}$, Zhongxue Gan$^{1}$, and Fei Gao$^{2, 3}$% <-this % stops a space
\thanks{This work was supported by Shanghai Municipal Science and Technology Major Project 2021SHZDZX0103, National Natural Science Foundation of China under Grant 62003299, the Shanghai Engineering Research Center of AI \& Robotics, Fudan University, China, and the Engineering Research Center of AI \& Robotics, Ministry of Education, China. (Corresponding author: Fei Gao, Zhongxue Gan.)}%
\thanks{$^{1}$Academy for Engineering and Technology, Fudan University, Shanghai, 200433, China. }%
\thanks{$^{2}$State Key Laboratory of Industrial Control Technology, Institute of Cyber-Systems and Control, Zhejiang University, Hangzhou, 310027, China.}%
\thanks{$^{3}$Huzhou Institute of Zhejiang University, Huzhou, 313000, China.}%
\thanks{Email: \{jlhou19, ganzhongxue\}@fudan.edu.cn; fgaoaa@zju.edu.cn}%
}

\begin{document}

\makeatletter
\let\@oldmaketitle\@maketitle% Store \@maketitle
\renewcommand{\@maketitle}{\@oldmaketitle% Update \@maketitle to insert...
	\captionsetup{type=figure}
	\centering
	\includegraphics[width=1.0\linewidth]
	{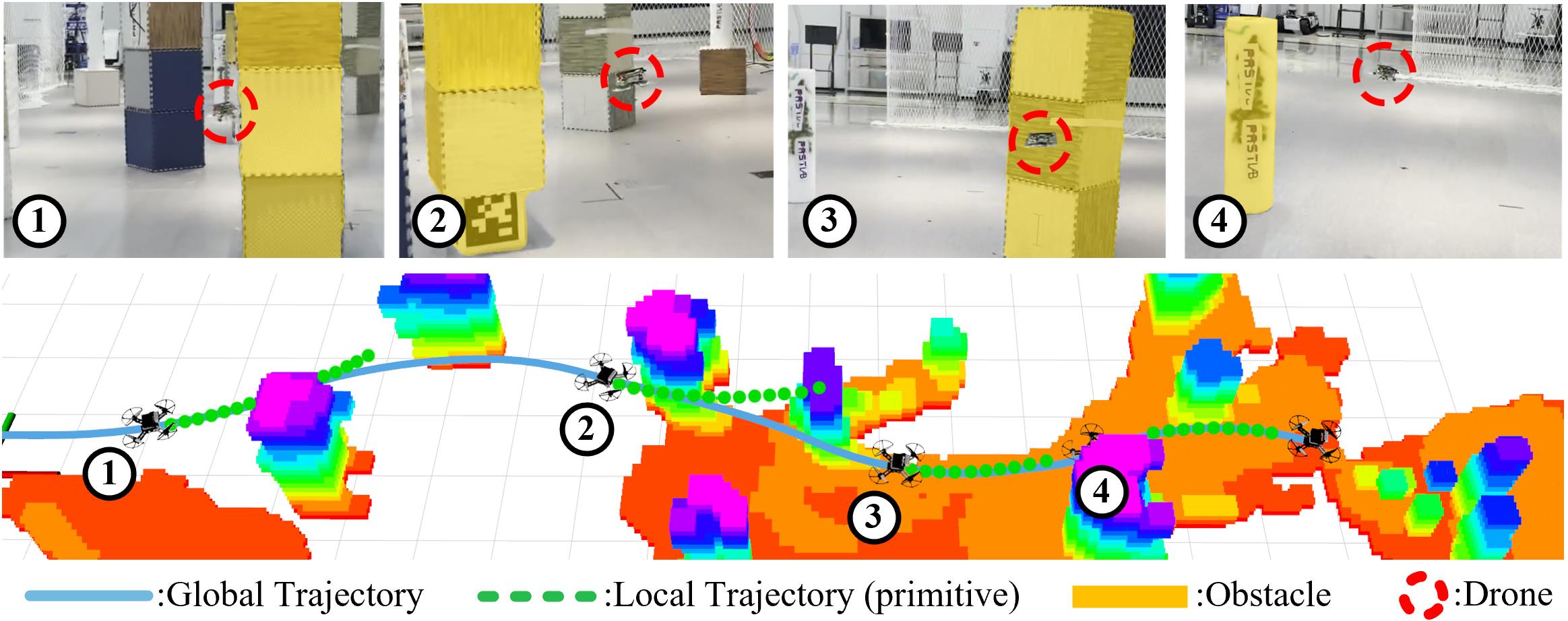}
	\label{pic:real_world}
	\captionof{figure}{
		Real-world experiment. The SWaP(Size, Weight, and Power) constrained quadrotor rapidly avoids obstacles one by one with a maximum expected speed of $2m/s$ using time-optimal primitives.
	}
}% ... an image
\vspace{-1.6cm}
\makeatother

\maketitle
\thispagestyle{empty}
\pagestyle{empty}

\begin{abstract}

It is a significant requirement for a quadrotor trajectory planner to simultaneously guarantee trajectory quality and system lightweight. Many researchers focus on this problem, but there's still a gap between their performance and our common wish. 
In this paper, we propose an ultra lightweight quadrotor planner with time-optimal primitives. 
Firstly, a novel motion primitive library is proposed to generate time-optimal and dynamical feasible trajectories offline.
Secondly, we propose a fast collision checking method with a deterministic time consumption, independent of the sampling resolution of the primitives. 
Finally, we select the minimum cost trajectory to execute among the safe primitives based on user-defined requirements.
The propsed transformation relation between the local trajectories ensures the smoothness of the global trajectory.
The planner reduces unnecessary online computing power consumption as much as possible, while ensuring a high-quality trajectory.
Benchmark comparisons show that our method can generate the shortest flight time and distance of trajectory with the lowest computation overload.
Challenging real-world experiments validate the robustness of our method.

\end{abstract}

%\textbf{\textit{Index Terms}--coordination}

%%%%%%%%%%%%%%%%%%%%%%%%%%%%%%%%%%%%%%%%%%%%%%%%%%%%%%%%%%%%%%%%%%%%%%%%%%%%%%%%

\section{INTRODUCTION}

In complex and unknown environments, trajectory planning plays a crucial role in autonomous robotic navigation.
It is a common wish that the onboard planner enjoys both efficiency and efficacy, especially for SWaP (Size, Weight, and Power) constrained robots like micro quadrotors.

Recently, our community has witnessed tremendous progress in quadrotor trajectory planning. An online trajectory planner usually consists of environment representation and trajectory generation.
For the former, researchers typically maintain a fused map from onboard sensors, such as the occupancy grid map, which requires a large computation overload to update \cite{zhou2020ego}. Some other methods\cite{ji2021mapless,florence2020integrated} are proposed using more lightweight data structures like kd-trees. However, the cost of rebuilding such data structures can still not be negligent, and the time complexity of querying collision status increases for the subsequent trajectory generation.
For the latter, most trajectory planners formulate and solve an optimization problem iteratively, generating a smooth, safe and dynamical feasible trajectory. However, these methods usually require further processing for environment representation, such as Euclidean Signed Distance Field (ESDF)\cite{zhou2020robust} or Safe Flight Corridor (SFC) \cite{liu2017planning}, which are more time-consuming.
Observing this shortcoming, the motion primitive methods\cite{ryll2019efficient,yang2021intention,florence2020integrated} generate a bunch of trajectories in a sampling space, and pick up a trajectory with the minimum user-defined cost after collision checking. 
Nevertheless, due to the limitation of the sampling resolution, the generated trajectories are far from optimum, sometimes even cannot find a feasible trajectory.

Investigating the above issues, we find that, the existing trajectory planners always trade off environment representation and trajectory generation, but cannot simultaneously decrease their overload under the guarantee of generating high-quality trajectories.
Based on such wish, we design an ultra lightweight planner with pre-computed time-optimal primitives. 
Firstly, we use time-optimal path parameterization on reachability analysis (TOPP-RA)\cite{pham2018new} to offline construct a motion primitive library, which contains time-optimal trajectories satisfying the dynamical constraints of a quadrotor. 
Secondly, we propose a fast collision checking method with deterministic time consumption.
Most conventional approaches check collision for each primitive on a maintained map one by one. 
In this way, the time consumption depends on the sampling resolution of the primitives.
Instead, we construct virtual voxel grids covering the motion primitives library, and pre-compute the spatial occupancy relationship of each grid with respect to the primitives.
When receiving sensor data like point clouds, the unsafe primitives are removed batch by batch according to which grid each point cloud locates in. Thus the time consumption is only related to the number of point clouds, and the sampling resolution of the primitives can be set sufficiently fine to provide near-optimal candidates.
%In this method, we pre-calculate the spatial occupancy of each motion primitive with respect to the attitude of a quadrotor. 
%While conducting collision status query, unlike traditional methods that check whether the sampled coordinates are occupied, we directly obtain the occupancy of trajectories through the spatial distribution of the point cloud.
Finally, we select the minimum cost trajectory among the safe primitives based on user-defined requirements such as smoothness and goal-approaching progress.
Our planner adopts the receding horizon planning strategy. In each (re)-planning, we establish a coordinate system $\{V\}$ with the velocity direction of the quadrotor as the x-axis, and transform the motion primitives into $\{V\}$. Since the pre-computed motion primitives are always tangent to the x-axis, the smoothness of the global trajectory is guaranteed.
%Our planner adopts the receding horizon replanning strategy, ensuring the smoothness of global trajectory by ingeniously designing the transformation relationship between local trajectories. 

We compare our method with two state-of-the-art lightweight quadrotor local planners\cite{ji2021mapless,zhou2022swarm}. 
The result shows that our method can not only generate trajectories with the shorter flight time and distance, but also require the lowest computational budget. 
Furthermore, extensive simulations and real-world experiments validate our proposed method on a customized SWaP quadrotor platform.
We will release our implementation as an open-source package\footnote{https://github.com/ZJU-FAST-Lab/Primitive-Planner}.
We summarize the contributions of this paper as follows:
\begin{enumerate}
	\item A novel motion primitive library, which adopts TOPP-RA\cite{pham2018new} method to offline generate time-optimal, dynamical feasible trajectories.
	\item A fast collision checking method, whose time complexity is independent of the sampling resolution of the primitives. It can also maintain safe clearance from obstacles by inflating the primitives offline.
	\item An ultra lightweight (re)-planning system composed of the above two efficient modules. The proposed planner quickly generates high-quality trajectories with lowest online overload. 
\end{enumerate}

\section{RELATED WORK}
\begin{comment}
\subsection{Quadrotor Trajectory Planning}

The classical planners formulate trajectory planning as a nonlinear optimization problem and online optimizes a feasible trajectory, which trades off motion smoothness, dynamical feasibility, and collision violation. However, these planners requires large computational power to extract free space or environment representation methods, 

such as occupancy grid map\cite{zhou2020ego}, ESDF\cite{zhou2020robust}, SFC\cite{liu2017planning}. Observing this shortcoming, our community has also been exploring more lightweight environment representation methods. Mapless\cite{ji2021mapless} kd-tree . Meanwhile, the motion primitive planners\cite{ryll2019efficient,florence2020integrated} generate a bunch of trajectories in a sampling space, use the kd-tree to select safe trajectories, and then pick up a trajectory with the minimum user-defined cost to execute. Because the sampled trajectories are not optimized and the sampling resolution is limited, the quality of the generated trajectory cannot always be guaranteed.
In summary, the above methods cannot realize the wish that enjoys both efficiency and efficacy.
\end{comment}

\subsection{Motion Primitive Library}

Motion primitive library in autonomous navigation usually generates multiple paths/trajectories using different sampling methods.
Zhang et al. \cite{zhang2020falco} propose a motion primitive library for offline sampling fixed-length paths in position space. However, the paths lack dynamical information, making it difficult to exert the robot's mobility.
Ryll et al. \cite{ryll2019efficient} sample different local end states and generate multiple fixed duration min-jerk trajectories in combination with the start state.
Yang et al. \cite{yang2021intention} sample different local end states in the velocity space $a=\{v_x, v_z, \omega\}$ online, then combining the start state to generate multiple fixed-duration 8-th order polynomial trajectories and integrating them to provide position information.
Nevertheless, many primitives generated by these two methods are dynamical-invalid. They require sampling a large number of primitives to find a dynamical feasible and safe trajectory.
Collins et al. \cite{collins2020efficient} adaptively sample the end states based on the start speed, which enhances the trajectory library's reliability but requires more computational power to select a feasible trajectory.
Bucki et al. \cite{bucki2020rectangular} propose a new pyramid partitioning method, which speeds up collision checking and saves computational power, but leads to more conservative trajectories and poor performance in obstacle-dense scenarios.
Florence et al. \cite{florence2020integrated} sample a set of constant control variables online and integrate them forward with a fixed duration to form a trajectory library.
The above methods aim to find a dynamical feasible and safe primitive, but without considering the time optimality of the trajectories.

\subsection{Time-optimal Path Parameterization (TOPP)}

TOPP is the problem of traversing a path in the fastest way possible while satisfying constraints. The numerical-based method \cite{pham2014general} directly solves the optimal control problem at each path position, which can be implemented quickly, but is less robust and difficult to deal with velocity bounds.
The optimization-based method \cite{verscheure2009time} robustly solves TOPP as a large convex optimization problem, which can satisfy the convex constraints, but with low solution efficiency.
TOPP-RA \cite{pham2018new} is a robust and efficient solver by solves some small LP problems.
TOPP-RA is mostly used for manipulators, and Ivanovic et al. \cite{ivanovic2022parabolic} use it for quadrotors, but it can only handle problems where the speed at the start and end of the path is zero. 

In this work, the proposed method can parameterize a non-zero start and end speed of the path based on TOPP-RA, which are realized on a quadrotor for the first time to the best of our knowledge.

%\section{Problem Formulation}

\section{System Overview}

\begin{figure}[t]
	\centering
	\includegraphics[width=1.0\linewidth]{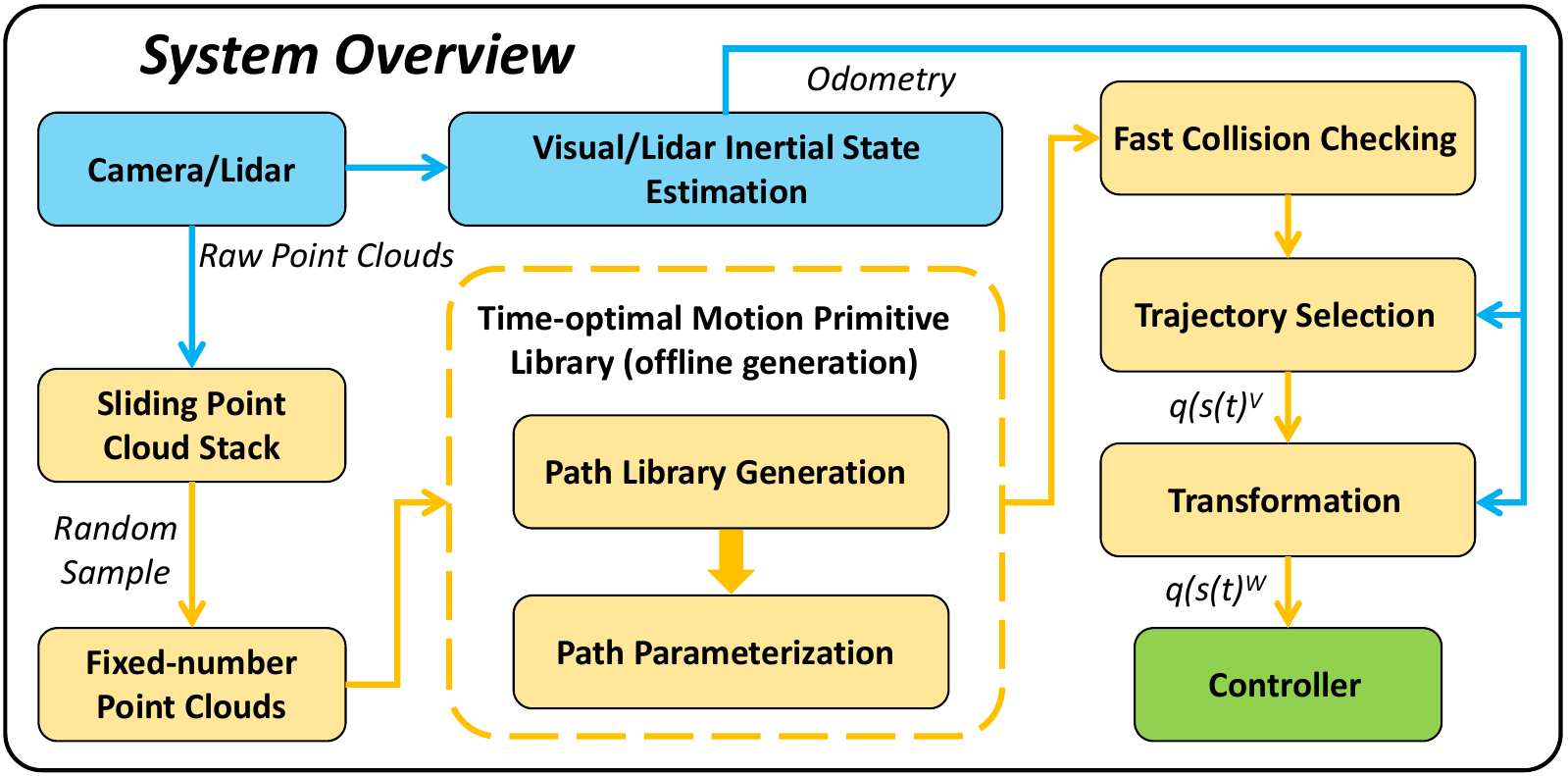}
	\captionsetup{font={small}}
	\caption{System overview.}
	\label{pic:system_overview}
	\vspace{-1.0cm}
\end{figure}
The proposed (re)-planning system (the yellow part) is shown in Fig. \ref{pic:system_overview}. It receives the output of camera/lidar and camera/lidar inertial state estimation (the blue part).
This (re)-planning system works in four steps. First, the time-optimal motion primitive library generates $N_{t}$ time-optimal trajectories offline (Sect.\ref{sec:mpl}).
Second, the sliding point cloud stack stores the raw point clouds of the latest $N_f$ frames. Fixed-number point clouds are sampled from the stack using the random sample method\cite{vitter1984faster} while ensuring environmental fidelity. The fast collision checking removes unsafe trajectories from the motion primitive library (Sect.\ref{sec:collision_check}).
Third, we select the minimum cost trajectory $q(s(t))^V$ by the user-defined requirement and the quadrotor's current speed (Sect.\ref{sec:traj_select}).
Lastly, the receding horizon planning obtains the trajectory $q(s(t))^W$ in the world coordinate system $\{W\}$ through the coordinate system transformation (Sect.\ref{sec:re_plan}). The controller executes the trajectory $q(s(t))^W$ (the green part).

\section{Methodology}

\subsection{Time-optimal Motion Primitive Library}
\label{sec:mpl}
We decouple the offline generation of the time-optimal motion primitive library into path library generation and path parameterization. Path library generation constructs a bunch of geometric paths (without dynamical information); Path parameterization determines time-optimal trajectories along the geometric paths, while considering dynamical constraints.

\subsubsection{Path Library Generation}
\label{subsubsec:lib}
\begin{figure}[t]
	\centering
	\includegraphics[width=1.0\linewidth]{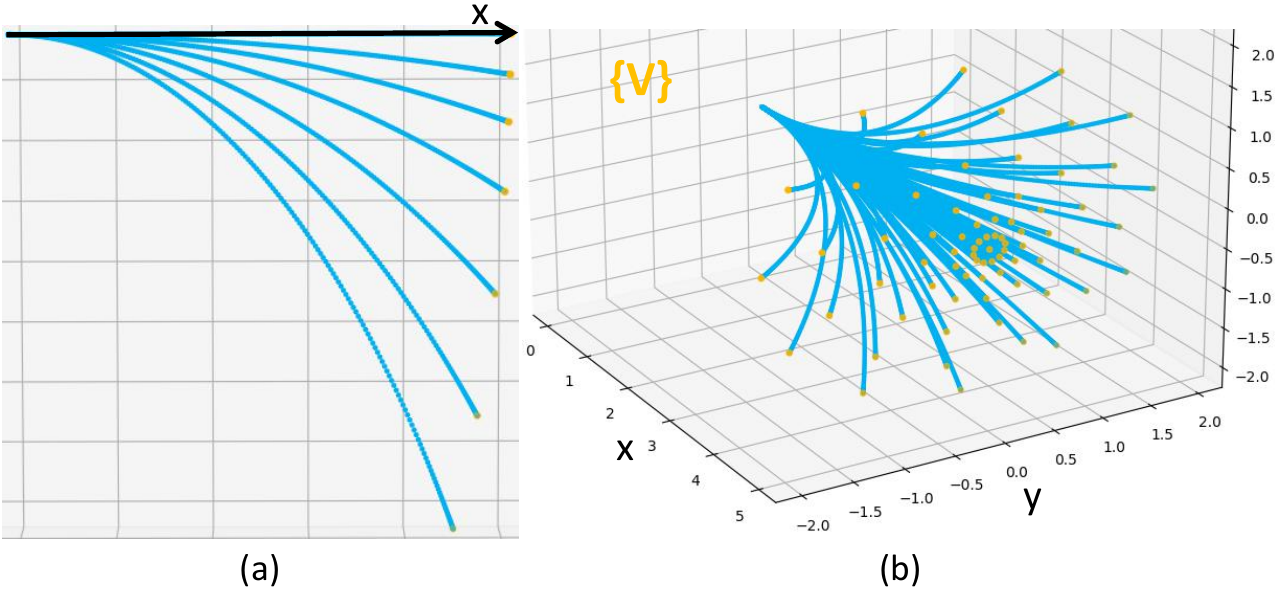}
	\captionsetup{font={small}}
	\caption{Path library generation; Blue splines are paths. The origin of all paths coincides with the origin of the coordinate system \{V\}. All paths are tangent to the x-axis. The orange dots are the end point of each path.}
	\label{pic:path_lib}
	\vspace{-0.4cm}
\end{figure}

% q(s)
In Fig. \ref{pic:path_lib}, we set 7 arcs with different radius $r = \{6, 8, 12, 20, 36, 78, \infty\}~m$ and a length of $5m$ (Fig. \ref{pic:path_lib}a). Then, these arcs are rotated at different start angles $\theta = \{0^\circ, -10^\circ, -20^\circ, 0^\circ, -10^\circ, -20^\circ, \backslash\}$ and interpolated by $30^\circ$ to obtain 73 paths (Fig. \ref{pic:path_lib}b).
Choosing different rotation start angles can increase the spatial distribution diversity of the path library, allowing it to fully cover the area where the quadrotor is about to travel.
We can also adjust $r$ and $\theta$ to generate different path libraries.

\subsubsection{Path Parameterization}
We adopt the robust and efficient time-optimal parameterization method TOPP-RA\cite{pham2018new} to implement path parameterization for non-zero start and end speed. Our method contains four steps, path discretization, constraint formulation, control set computation, time-optimal trajectory generation, as illustrated in Algorithm 1.

\textbf{Path Discretization}: We adopt a high-accuracy interpolation scheme\cite{pham2018new} to uniformly discretize a path $q(s) \in \mathbb{R}^n ~(\textup{path parameter}~s:[0, 1] \rightarrow [s_0, s_N]$) into $N+1$ discrete points with interval $\Delta$
\begin{equation}
s_0, s_1, ... s_{N-1}, s_N.
\end{equation} 
where $\Delta = (s_N - s_0) / N$.

The velocity and acceleration of $i-th$ discrete point are respectively $\dot{s}_i, \ddot{s}_i$ in the interval $[s_i, s_{i+1}]$, with the following relation:
\begin{equation}
\dot{s}_{i+1}^2 = \dot{s}_i^2 + 2 \Delta \ddot{s}_i.
\end{equation} 

\textbf{Constraint Formulation} :We construct constraints at each discrete point on the path. The general second-order constraint form \cite{hauser2014fast,pham2018new} is as follows

\begin{equation}
\label{eq:sec_constraint}
A(q)\ddot{q} + \dot{q}^TB(q)\dot{q} + f(q) \in \mathscr{C}(q),
\end{equation} 
where $A:\mathbb{R}^n \rightarrow \mathbb{R}^{m \times n}$, $B:\mathbb{R}^n \rightarrow \mathbb{R}^{n \times m \times n}$, $f:\mathbb{R}^n \rightarrow \mathbb{R}^{m}$. $\mathscr{C}(q) \in \mathbb{R}^{m}$ is the constraint. 

Differentiating path $q(s)$ has
\begin{gather}
\label{eq:diff}
\dot{q} = q'\dot{s},~~~\ddot{q} = q''\dot{q}^2 + q'\ddot{s}
\end{gather}
where $'$ denotes the differential with respect to the path parameter $s$, $\dot{}$ denotes the differential with respect to time.

Substituting Eq. \ref{eq:diff} into Eq. \ref{eq:sec_constraint} has
\begin{equation}
\label{eq:subs_constraint}
a(s)\ddot{s} + b(s)\dot{s}^2 + c(s) \in \mathscr{C}(s).
\end{equation} 
where
\begin{subequations}
\label{eq:subs_coeff}
\begin{gather}
a(s) := A(q(s))q'(s),\\
b(s) := A(q(s))q''(s) + q'(s)^TB(q(s))q'(s),\\
c(s) := f(q(s)),\\
\mathscr{C}(s) := \mathscr{C}(q(s)).
\end{gather}
\end{subequations}

Since we set the optimization variable to be $\dot{s}^2, \ddot{s}$ in linear programming (LP) problem, the linear constraint is constructed as
\begin{equation}
\label{eq:inequ_con}
F\mathscr{C}(s) \leq g,
\end{equation}
\begin{equation}
\label{eq:vel_square}
\dot{s}^2_{min} \leq \dot{s}^2 \leq \dot{s}^2_{max},
\end{equation}
\begin{equation}
\ddot{s}_{min} \leq \ddot{s} \leq \ddot{s}_{max},
\end{equation}
where $F \in \mathbb{R}^{n \times m}$ is a constant matrix.

According to the quadrotor's velocity bound $[v_{min}, v_{max}]$, acceleration bound $[a_{min}, a_{max}]$, and maximum velocity norm $v_{norm}$, the corresponding velocity constraint is as follows
\begin{equation}
\label{eq:rob_vel}
v_{min} \leq \dot{q} \leq v_{max},
\end{equation}
substituting Eq. \ref{eq:diff}, Eq. \ref{eq:vel_square} into Eq. \ref{eq:rob_vel} has
\begin{gather}
\dot{s}^2_{min} = 0,~~~\dot{s}^2_{max} = v_{max}^2 / (q'^Tq').
\end{gather}
The corresponding acceleration constraint is as follows
\begin{equation}
\label{eq:rob_acc}
a_{min} \leq \ddot{q} \leq a_{max},
\end{equation}
substituting Eq. \ref{eq:diff}, Eq. \ref{eq:subs_constraint}, Eq. \ref{eq:inequ_con} into Eq. \ref{eq:rob_acc} has
\begin{equation}
\begin{split}
a(s) = q'(s),~b(s) = q''(s),\\
F=[\mathbf{I},-\mathbf{I}]^T,~g = [a^T_{max}, -a^T_{min}]^T.
\end{split}
\end{equation}
The corresponding velocity norm constraint is as follows
\begin{equation}
\label{eq:rob_norm}
\dot{q}^2 \leq v_{norm}^2,
\end{equation}
substituting Eq. \ref{eq:diff}, Eq. \ref{eq:subs_constraint}, Eq. \ref{eq:inequ_con} into Eq. \ref{eq:rob_norm} has
\begin{gather}
b(s) = q'(s)^Tq'(s),~F=\mathbf{I},~g = v_{norm}^2.
\end{gather}

\begin{algorithm}[t]
	\label{alg:topp_ra}
	
	\KwIn{Path $q(s)$\\
		quadrotor's start and end point velocities $\dot{q(s)}_0, \dot{q(s)}_N$\\
		velocity bound $[v_{min}, v_{max}]$\\
		acceleration bound $[a_{min}, a_{max}]$\\
		velocity norm $v_{norm}$\\}
	\KwOut{Time-optimal trajectory $q(s(t))$}
	
	/*  \textbf{Path Discretization}*/\\
	$\{s_i\}_{0:N}$ = UniformDiscretize($q(s)$)\\
	/*  \textbf{Constraint Formulation}*/\\
	pc\_vel = VelConstraintParameter($[v_{min}, v_{max}]$)\\
	pc\_acc = AccConstraintParameter($[a_{min}, a_{max}]$)\\
	pc\_norm = NormConstraintParameter($v_{norm}$)\\
	instance = TOPPRA(pc\_vel, pc\_acc, pc\_norm, $q(s)$)\\
	%$q(s(t))$ = instance.\textbf{compute\_trajectory}($\dot{q}_0, \dot{q}_N$)\\
	/*  \textbf{Control Set Computation}*/\\
	$K$ = ComputeControlSets(instance, $\dot{s}_N^2$)\\
	/*  \textbf{Time-optimal Trajectory Generation}*/\\
	\ForEach{$i = 0:N-1$}{
		/* Compute optimization variables */\\
		$[\dot{s}^2_i,~\ddot{s}_i]$ = Forward($\dot{s}_i$, $K_{i+1}$)\\
		%$\dot{s}_{i+1}^2 = \dot{s}_{i}^2 + 2 * (s_{i+1} - s_{i}) * \ddot{s}_i$
	}
	\ForEach{$i = 1:N$}{
		$\dot{s}_{average} = (\dot{s}_{i-1} + \dot{s}_i) / 2$\\
		$\Delta t = (s_{i} - s_{i-1}) / \dot{s}_{average}$\\
		$t_i = t_{i-1} + \Delta t$ 
	}
	$q(s(t))$ = CubicSpline($q(s)$, $t$)
	
	\textbf{return} $q(s(t))$\\	
	\caption{Path Parameterization with TOPP-RA}
\end{algorithm}
\vspace{-0.0cm}

\textbf{Control Set Computation}: According to the reachability analysis, the feasible speed square interval $\mathbb{I}_{s_i}$ at each discrete point $s_i$ is calculated sequentially from the speed square of the end point on the path $\dot{s}^2_N$ (backward).
For example, if $\mathbb{I}_{s_{i+1}}$ is known, we can use the following LP problem to solve $\mathbb{I}_{s_i}$:
\begin{subequations}
	\label{eq:backward}
	\begin{gather}
	\mathbb{I}^-_{s_i} = \min_{\dot{s}^2, \ddot{s}}~\dot{s}^2_i,~~~\mathbb{I}^+_{s_i} = \max_{\dot{s}^2, \ddot{s}}~\dot{s}^2_i,\\
	\textup{s.t.}~v_{min} \leq \dot{q} \leq v_{max},\\
	a_{min} \leq \ddot{q} \leq a_{max},\\
	\dot{q}^2 \leq v_{norm}^2,\\
	\dot{s}^2_i+2 \Delta \ddot{s}_i \in \mathbb{I}_{s_{i+1}}.
	\end{gather}
\end{subequations}

\begin{comment}
\begin{subequations}
	\label{eq:backward}
	\begin{align}
	\mathbb{I}^-_{s_i} =& \min_{\dot{s}^2, \ddot{s}}~\dot{s}^2_i,~~~\mathbb{I}^+_{s_i} = \max_{\dot{s}^2, \ddot{s}}~\dot{s}^2_i,\\
	\textup{s.t.}~&v_{min} \leq \dot{q} \leq v_{max},\\
	&a_{min} \leq \ddot{q} \leq a_{max},\\
	&\dot{q}^2 \leq v_{norm}^2,\\
	&\dot{s}^2_i+2 \Delta \ddot{s}_i \in \mathbb{I}_{s_{i+1}}.
	\end{align}
\end{subequations}
\end{comment}
The set of velocity square intervals at all discrete points is defined as the control set $K$.

\textbf{Time-optimal Trajectory Generation}: We start with the computation of maximum feasible speed square $\dot{s}^2_{i~max}$.  At each discrete point,  $\dot{s}^2_{i~max}$  is determined sequentially from the speed square of the start point $\dot{s}^2_0$ on the path (forward). For example, if $\dot{s}^2_{i-1}$ is known, we can use the following LP problem to solve $\dot{s}^2_{i~max}$:
\begin{subequations}
	\label{eq:forward}
	\begin{align}
	\dot{s}^2_{i~max} =& \max_{\dot{s}^2, \ddot{s}}~\dot{s}^2_{i-1} + 2\Delta \ddot{s}_{i-1}\\
	\textup{s.t.}~&v_{min} \leq \dot{q} \leq v_{max},\\
	&a_{min} \leq \ddot{q} \leq a_{max},\\
	&\dot{q}^2 \leq v_{norm}^2\\
	&\dot{s}^2_i \in K_i.
	\end{align}
\end{subequations}

The optimal set of time $t$ is calculated through the average speed using all maximum feasible speeds. Then parameterizing the path $q(s)$ obtains the time-optimal trajectory $q(s(t))$.

The interpolation scheme has a satisfaction error of $O(\Delta^2)$ \cite{pham2018new}, which is related to the discrete point number $N_{dp}$. We set $N_{dp}=1000$ to tradeoff efficiency and exactness. We utilize the Seidel's algorithm\cite{seidel1991small} to solve Eq. \ref{eq:backward} and Eq. \ref{eq:forward}.

%Because the motion primitive library is generated offline, the quality of the optimization is more important than the computation time. Thus, we use the Seidel's LP algorithm \cite{seidel1991small} which can get an exact solution.
%We use the Seidel's LP algorithm \cite{seidel1991small} which can get an exact solution.
%Since the continuous-time constraint cannot be added to the constrained optimization problem, the above optimization problems use a high-accuracy interpolation scheme to discretize each path and impose dynamical constraints on the discrete points. 
%The interpolation scheme has a satisfaction error of $O(\Delta^2)$. 
%Selecting the number of discrete points $N=1000$ can ensure the solution's accuracy. 
%For more details about satisfaction errors, please refer to \cite{pham2018new}.

%In summary, we generate a path library, discretize all paths, and parameterize the paths to obtain a time-optimal motion primitive library under the premise of satisfying dynamical constraints. As a result, the motion primitive library has good spatial coverage, considers time optimality, and satisfies the robot's dynamical constraints.

\begin{figure}[t]
	\centering
	\includegraphics[width=1.0\linewidth]{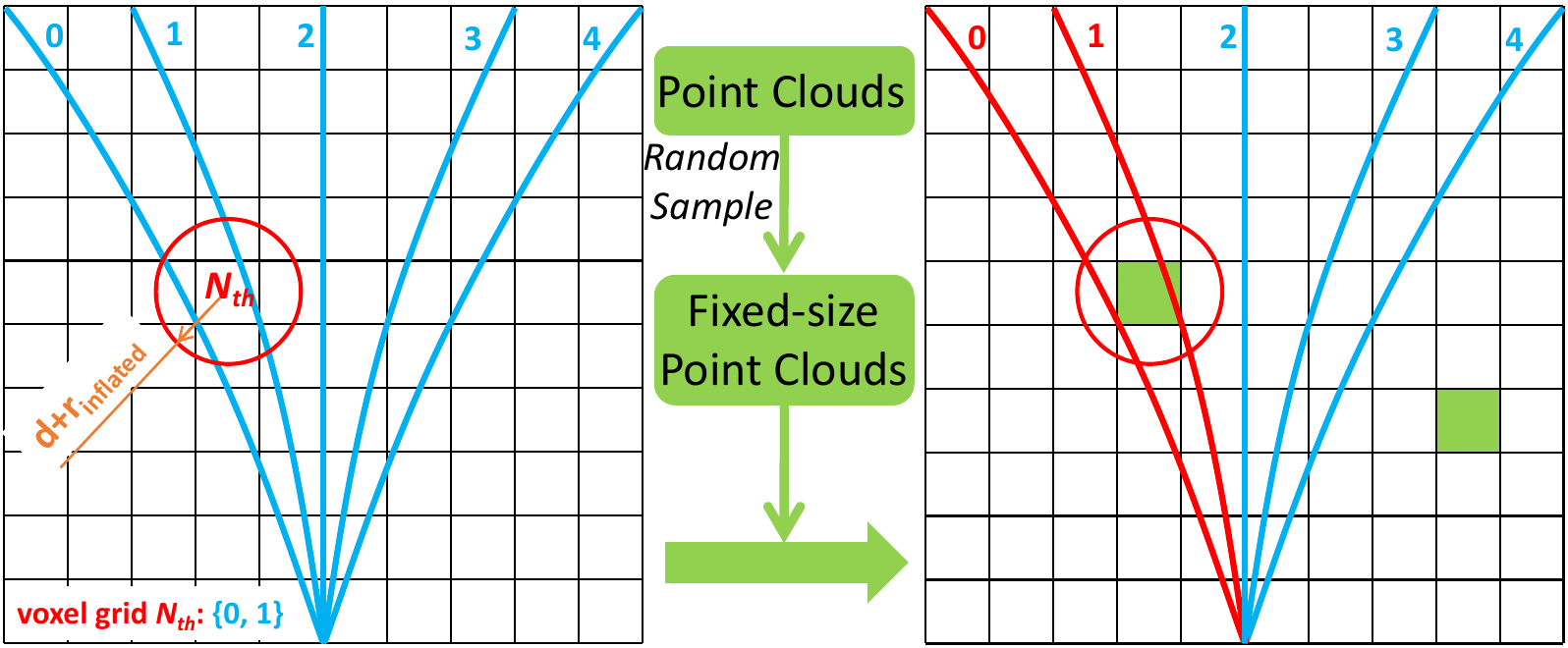}
	\captionsetup{font={small}}
	\caption{Fast collision checking; Voxel grid $N_{th}$ associates paths $0$ and $1$. The blue paths are free. The red paths are occupied.}
	\label{pic:collision_check}
	\vspace{-0.8cm}
\end{figure}

For each sampled path in Sec.\ref{subsubsec:lib},  we set the same end velocity $\dot{q(s)}_N = 0$ and change the start velocity $\dot{q(s)}_{0} = [0, 0.1, ... v_{max}]~m/s$ to generate the motion primitive library. The motion primitive library has good spatial coverage, considers time optimality, and satisfies the quadrotor's dynamical constraints.

\subsection{Fast Collision Checking}
\label{sec:collision_check}

To ensure safe navigation of quadrotors in unknown environments, fast collision checking is necessary.
Typical collision checking methods \cite{ryll2019efficient,collins2020efficient} fuse the obstacles' point clouds into a gridmap or build a kd-tree online, and then discretely sample each primitive to query whether a collision occurs.
These methods have indeterminate collision checking time depending on the sampling resolution of the primitives.
Inspired by Zhang et al. \cite{zhang2020falco}, we establish a fast collision checking method with deterministic time, whose time complexity is independent of the sampling resolution of the primitives. It can also maintain safe clearance from obstacles by inflating the primitives offline.

In Fig. \ref{pic:collision_check}, we offline construct all the primitives in the motion primitive library as a kd-tree, and split the local coverage space of the motion primitives into fixed-size virtual voxel grids.
The spatial occupancy relationship of each grid with respect to the primitives can be pre-computed by querying the kd-tree within a specific distance $d$ from the center of each grid.
We use the online random sampling method \cite{vitter1984faster} to obtain fixed-number $N_{pc}$ point clouds from the sliding stack under the premise of ensuring the fidelity of environment. 
The unsafe primitives are removed batch by batch according to which grids the fixed-number point clouds locate in.
Thus the time consumption is only related to the number of point clouds, and is independent of the sampling resolution of the primitives.
This proposed method reduces the number of sampled point clouds to a fixed number $N_{pc}$, so that collision checking can be finished quickly in deterministic time.
To ensure the quadrotor's safety, it is usually necessary to maintain safe clearance from obstacles. This method only needs to change the query distance $d$ to $d+r_{inflated}$ ($r_{inflated}$ is the inflated radius).

\subsection{Trajectory Selection}
\label{sec:traj_select}
\begin{figure}[t]
	\centering
	\includegraphics[width=1.0\linewidth]{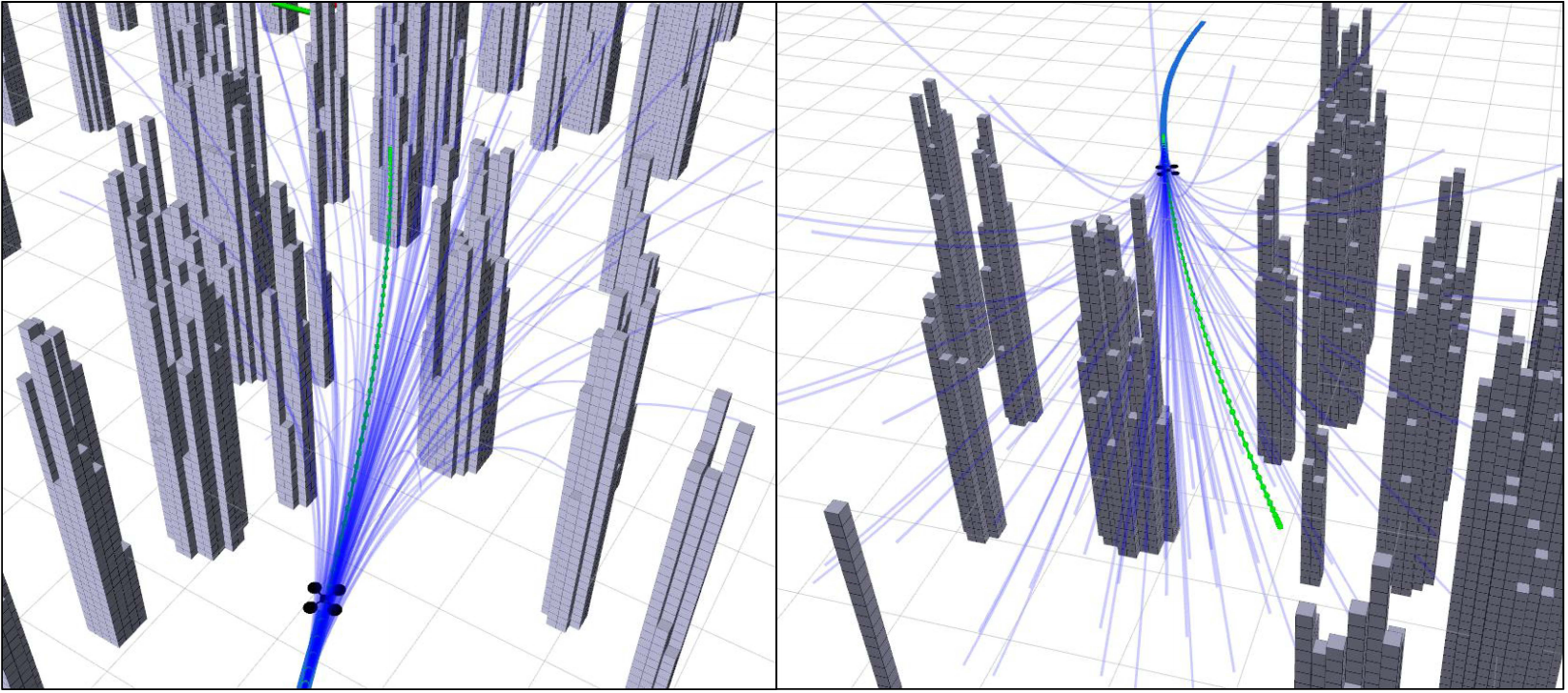}
	\captionsetup{font={small}}
	\caption{Trajectory selection; The blue curves represent all trajectories in the motion primitive library. The green curve represents the selected trajectory.}
	\label{pic:traj_select}
	\vspace{-1.2cm}
\end{figure}
We first exclude unsafe trajectories from the collision checking (To ensure safety, if all trajectories in the motion primitive library are occupied, the quadrotor executes an emergency stop command). Then, we construct the cost function $c$ of each trajectory representing the user-defined requirements in Equ \ref{eq:cost}. 
The goal cost $c_{goal}$ represents the end position of the trajectory $p_{end}$ as close as possible to the global goal $p_{goal}$. The boundary cost $c_{bound}$ represents the end position $p_{end}$ cannot be outside of the allowed bounds $B$.
Other desired requirements can be implemented by extending the cost function $c$. Finally, we select the minimum cost trajectory with the start speed equal to the quadrotor's current speed in Fig. \ref{pic:traj_select}.
\begin{subequations}
	\label{eq:cost}
	\begin{gather}
	c = \lambda_g c_{goal} + \lambda_b c_{bound}\\
	c_{goal} = \lVert p_{end} - p_{goal} \rVert - \lVert p_{start} - p_{goal} \rVert\\
	c_{bound} = \begin{cases}
	\textup{constant}, & \text{if}~p_{end} \notin B\\
	0, & \text{if}~p_{end} \in B
	\end{cases}
	\end{gather}
\end{subequations}
where $\lambda_g, \lambda_b$ are the corresponding weights. $p_{start}$ indicates the current quadrotor's position.

\subsection{Receding Horizon Planning}
\label{sec:re_plan}
We adopt the receding horizon planning strategy, and the (re)-planning frequency is $10$ Hz.
In each (re)-planning, the quadrotor's current state is used as the next trajectory start state, so the trajectory consistent with the quadrotor's current velocity $\textbf{v}_{current}$ is selected.
We make the trajectory tangent to the quadrotor's current velocity direction at the start point. The transformation relationship of the trajectory $T_{WV}$ from the velocity coordinate system $\{V\}$ to the world coordinate system $\{W\}$ is as follows
\begin{subequations}
\begin{gather}
\textbf{x}_{axis} = \textbf{v}_{current} / \lVert \textbf{v}_{current} \rVert,\\
\textbf{y}_{axis} = \textbf{x}_{axis} \times (0, 0, -1)^T,\\
\textbf{z}_{axis} = \textbf{x}_{axis} \times \textbf{y}_{axis},\\
R_{WV} = [\textbf{x}_{axis}, \textbf{y}_{axis}, \textbf{z}_{axis}],\\
T_{WV} = [R_{WV} ; p_{start}].
\end{gather}
\end{subequations}

\section{Evaluation}

In this section, we conduct detailed evaluation tests on the proposed contributions. All programs run on an Intel Core i7-10700 2.90GHz CPU.
All tests are performed on the $26\times20m$ random maps, where each obstacle is an average radius $r_{obs}=0.6m$ cylinder, as shown in Fig. \ref{pic:comparison}.
In each test, the start and end positions of the quadrotor are set to $[-18.0, -9.0, 1.0]~m$ and $[18.0, 9.0, 1.0]~m$, respectively.
The maximum speed and acceleration are set to $3m/s$ and $6m/s^2$. All comparative tests use the controlled variable method.

\subsection{Performance Analysis of Primitive-Planner}
\label{sec:per_analysis}

\begin{figure}[t]
	\centering
	\includegraphics[width=1.0\linewidth]{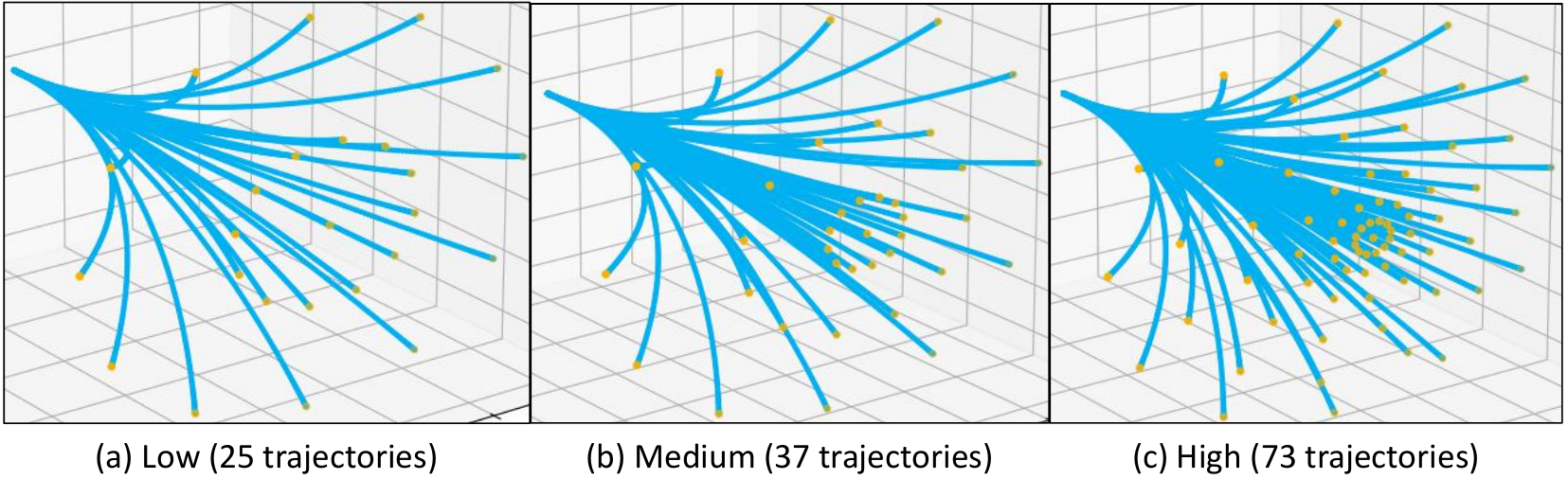}
	\captionsetup{font={small}}
	\caption{Three motion primitive libraries; (a) $r=\{8, 20, \infty\}m$ (b) $r=\{6, 12, 36, \infty\}m$ (c) $r = \{6, 8, 12, 20, 36, 78, \infty\}m$}
	\label{pic:three_motion_lib}
	\vspace{-1.5cm}
\end{figure}

\begin{table}[h]
	\centering
	\caption{Success Rate (\%) in Different Situations}
	\renewcommand{\arraystretch}{1.50}
	\label{tab:success_rate}
	\begin{tabular}{|c|c|c|c|}
		\hline
		\textbf{\diagbox{\# Trajectories}{\# Obstacles}} & \textbf{\begin{tabular}[c]{@{}c@{}}Sparse\\ (100)\end{tabular}} & \textbf{\begin{tabular}[c]{@{}c@{}}Medium\\ (150)\end{tabular}} & \textbf{\begin{tabular}[c]{@{}c@{}}Dense\\ (200)\end{tabular}} \\ \hline
		\textbf{Low (25)}    & 100     & 80     & 65    \\ \hline
		\textbf{Medium (37)} & 100    & 90     & 75    \\ \hline
		\textbf{High (73)}   & 100    & 100    & 100   \\ \hline
	\end{tabular}
	\vspace{-0.2cm}
\end{table}

We set up three motion primitive libraries (in Fig. \ref{pic:three_motion_lib}) with the trajectories' number $N_{t}=\{25, 37, 73\}$ and three scenes with obstacles' number $N_{obs}=\{100, 150, 200\}$.
The quadrotor runs $20$ times from the start to the end position in each situation (9 situations in total), with the success rate in Table \ref{tab:success_rate}.
The motion primitive library with low trajectories' number performs poorly in dense environments.
By increasing the trajectories' number, the motion primitive library can handle complex dense environments. 
Therefore, the proposed method is competent for autonomous navigation tasks in complex dense environments.

\subsection{Deterministic Time Collision Checking}

\begin{figure}[t]
	\centering
	\includegraphics[width=1.0\linewidth]{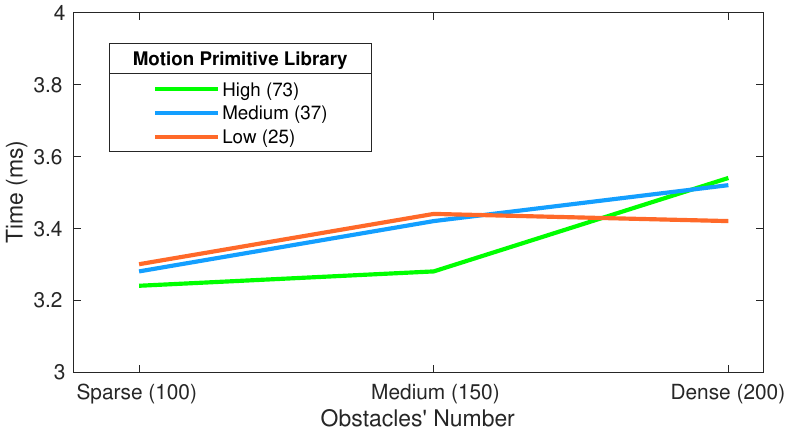}
	\captionsetup{font={small}}
	\caption{Collision checking time.}
	\label{pic:collision_time}
	\vspace{-1.5cm}
\end{figure}

We test the collision checking time of the three motion primitive libraries under different obstacle density environments.
We set the fixed-number of point clouds $N_{pc} = 2000$, which can ensure the fidelity of the environment.
Each situation runs $20$ times. The results are shown in Fig. \ref{pic:collision_time}.
As the number of motion primitive increases, the collision checking time does not vary significantly and stays within the $3.2-3.6~ms$ level.
When the number of sampled point clouds is less than 2000, the collision check time will decrease. This is also why the time varies slightly in different situations.
The results show that the proposed method can quickly check for collisions in deterministic time. And the time complexity is independent of the sampling resolution of the
primitives.

\begin{comment}
% \usepackage{multirow}
\begin{table}[h]
	\centering
	\caption{Collision Check Time (ms)}
    \renewcommand{\arraystretch}{1.50}
	\label{tab:colliison_time}
	\begin{tabular}{|cc|c|c|c|}
		\hline
		\multicolumn{2}{|c|}{\textbf{Trajectories' Number}}                        & \textbf{Low (25)} & \textbf{Medium (37)} & \textbf{High (73)} \\ \hline
		\multicolumn{1}{|c|}{\multirow{3}{*}{\textbf{Time (ms)}}} & \textbf{Min} & 3.04         & 3.28            & 3.25          \\ \cline{2-5} 
		\multicolumn{1}{|c|}{} & \textbf{Avg} & 4.07 & 4.16 & 4.11 \\ \cline{2-5} 
		\multicolumn{1}{|c|}{} & \textbf{Max} & 5.39 & 5.81 & 5.67 \\ \hline
	\end{tabular}
	\vspace{-0.4cm}
\end{table}
\end{comment}

\begin{table}[h]
	\centering
	\caption{Benchmark Comparison; $t_{total}$ and $d_{total}$ indicate the flight time and distance. $t_{env}$ indicates the environment representation time. $t_{com}$ indicates the average time to generate a local trajectory.}
	\renewcommand{\arraystretch}{1.50}
	\label{tab:comparison}
	\resizebox{\columnwidth}{!}{%
		\begin{tabular}{|c|c|c|c|c|}
			\hline
			\textbf{Method}   & \textbf{$t_{total}$ (s)} & \textbf{$d_{total}$ (m)} & \textbf{$t_{env}$ (ms)} & \textbf{$t_{com}$ (ms)} \\ \hline
			\textbf{Mapless}        & 19.747 & 48.541 & 2.567 & 8.227 \\ \hline
			\textbf{EGO-Planner-v2} & 21.707 & 41.934 & 9.233 & 0.741 \\ \hline
			\textbf{Proposed} & \textbf{13.479}     & \textbf{41.336}     & \textbf{/}          & 3.54 + 0.055        \\ \hline
		\end{tabular}%
	}
	\vspace{-0.6cm}
\end{table}

\subsection{Benchmark Comparisons}

\begin{figure}[t]
	\centering
	\includegraphics[width=1.0\linewidth]{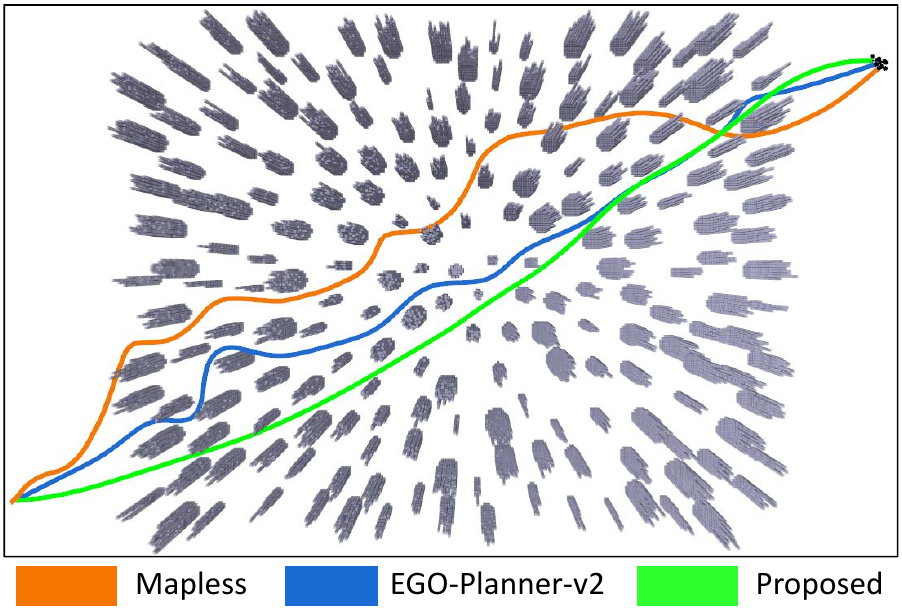}
	\captionsetup{font={small}}
	\caption{Trajectory visualization.}
	\label{pic:comparison}
	\vspace{-0.1cm}
\end{figure}

\begin{figure}[t]
	\centering
	\includegraphics[width=0.9\linewidth]{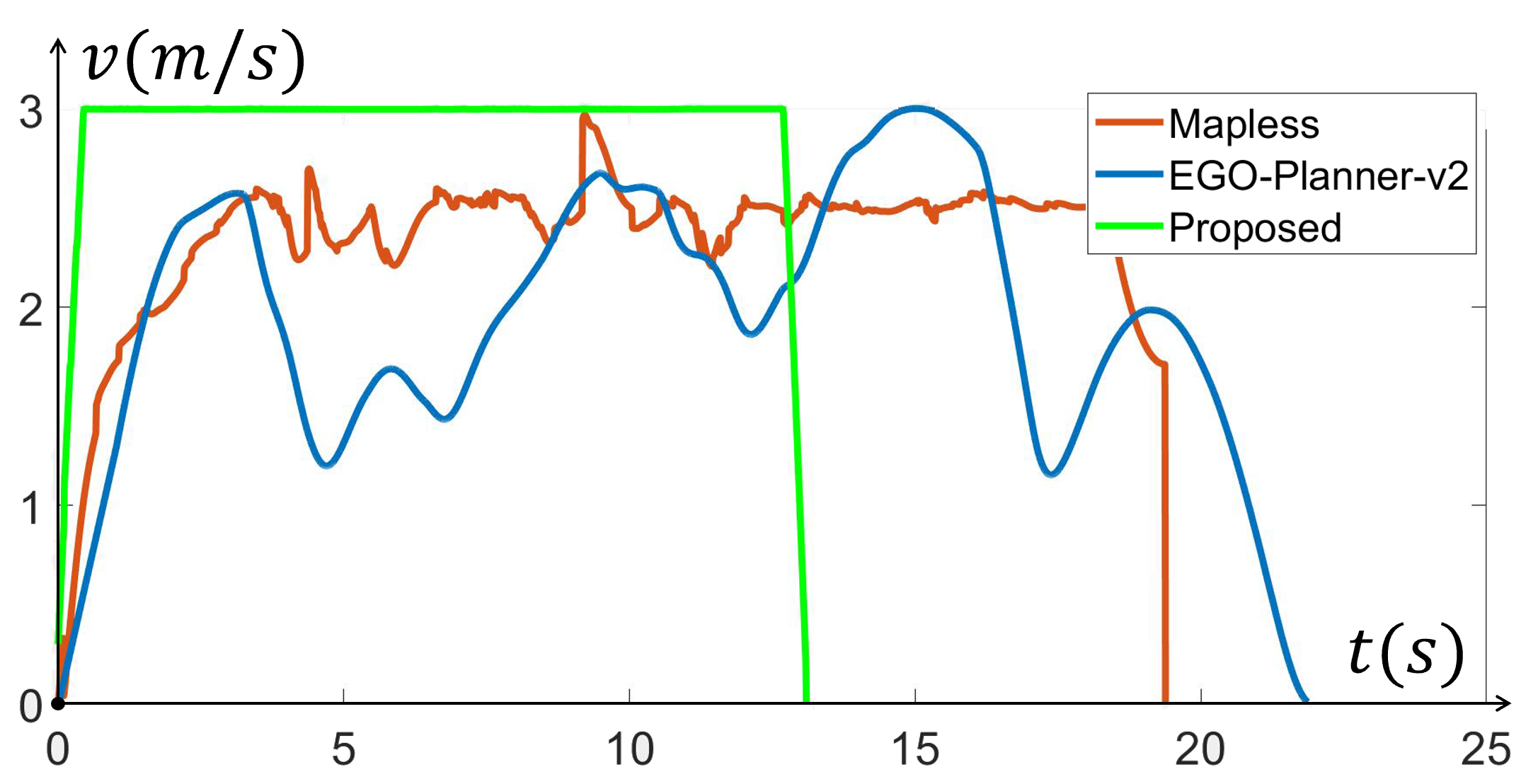}
	\captionsetup{font={small}}
	\caption{Velocity comparison.}
	\label{pic:vel_curve}
	\vspace{-1.5cm}
\end{figure}

In TABLE \ref{tab:comparison}, we compare the proposed method with two state-of-the-art methods, Mapless \cite{ji2021mapless} and EGO-Planner-v2 \cite{zhou2022swarm}.
Mapless is a lightweight approach with the kd-tree data structure. EGO-Planner-v2 is an online optimization method with a fusion map (the upgraded version of EGO-Planner \cite{zhou2020ego}).
We set the trajectories' number $N_{t} = 73$ in our method.
In dense environments ($N_{obs}=200$), each method runs 10 times from start to end position. All data is the average of $10$ experiments. The flight distance is calculated by integrating the position, which is slightly less than the actual distance.
The trajectories of the three methods in a random dense map are shown in Fig \ref{pic:comparison}. Their velocity profiles are shown in Fig. \ref{pic:vel_curve}.

In comparison, the proposed method has top performance with the shortest flight time, distance, and computation overload ($t_{env} + t_{com}$). The online computation overload of the proposed method only has $t_{com}$, which contains collision checking time ($3.54ms$) and trajectory selection time ($0.055ms$). 
This is mainly attributed to offline generated time-optimal motion primitive library and ultra lightweight (re)-planning system.

\section{REAL-WORLD EXPERIMENTS}
To validate the practical performance of the proposed method, we deploy it in a SWaP-constrained quadrotor platform whose configuration is detailed in \cite{zhou2022swarm}.
We offline generate the motion primitive library with different radius $r=\{2,3,4,6,8,12,20,36,78,\infty\}m$ and a length of $3m$, which contains $109$ time-optimal trajectories. We set the quadrotor speed and acceleration limits to $2m/s$ and $6m/s^2$, respectively. In Fig. 1, we demonstrate the safe autonomous navigation of the quadrotor in a dense environment. The quadrotor rapidly avoids obstacles one by one with a speed of almost 2m/s, validating the robustness of the proposed method. The smooth global trajectory validates the reliability of the transformation relation during (re)-planning.

\section{CONCLUSIONS AND FUTURE WORK}

In this paper, we propose an ultra lightweight quadrotor planner with time-optimal primitives. The planner reduces unnecessary online computing power consumption as much as possible, while ensuring a high-quality trajectory. Extensive simulations and challenging real-world experiments fully validate the high performance and robustness of the proposed method. In the future, the online overload of collision checking will be further optimized. We will also explore the application of the proposed method in the field of transportation.

\addtolength{\textheight}{-12cm}   % This command serves to balance the column lengths

%%%%%%%%%%%%%%%%%%%%%%%%%%%%%%%%%%%%%%%%%%%%%%%%%%%%%%%%%%%%%%%%%%%%%%%%%%%%%%%%

\bibliographystyle{ieeetr}
\bibliography{ICRA2023hjl}

\end{document}